\documentclass{article}

\usepackage{arxiv}

\usepackage[utf8]{inputenc} 
\usepackage[T1]{fontenc}    
\usepackage{hyperref}       
\usepackage{url}            
\usepackage{booktabs}       
\usepackage{amsfonts}       
\usepackage{nicefrac}       
\usepackage{microtype}      
\usepackage{lipsum}
\usepackage{times}
\usepackage{epsfig}
\usepackage{graphicx}
\usepackage{amsmath}
\usepackage{amssymb}
\usepackage{rotating}
\usepackage{multirow}
\usepackage{subfig}
\usepackage{bm}
\usepackage{hhline}
\usepackage[norule,symbol,perpage]{footmisc}
\DeclareUnicodeCharacter{FB01}{fi}
\title{LivDet 2021 Fingerprint Liveness Detection Competition - Into the unknown}

\author{Roberto Casula, Marco Micheletto, Giulia Orrù, Rita Delussu, Sara Concas, Andrea Panzino, Gian Luca Marcialis\\
University of Cagliari, Piazza d'Armi, I-09123 Italy\\
\\
{\tt\small \{roberto.casula,marco.micheletto,giulia.orru,rita.delussu,marcialis\}@unica.it}
}
\usepackage{tikz}
\usepackage{textcomp}
\usepackage{hyperref}

\newcommand\copyrighttext{%
  \footnotesize \textcopyright 2021 IEEE. Personal use of this material is permitted.
  Permission from IEEE must be obtained for all other uses, in any current or future 
  media, including reprinting/republishing this material for advertising or promotional 
  purposes, creating new collective works, for resale or redistribution to servers or 
  lists, or reuse of any copyrighted component of this work in other works. 
  DOI: \href{<https://ieeexplore.ieee.org/abstract/document/9484399>}{10.1109/IJCB52358.2021.9484399}}
\newcommand\copyrightnotice{%
\begin{tikzpicture}[remember picture,overlay]
\node[anchor=south,yshift=10pt] at (current page.south) {\fbox{\parbox{\dimexpr\textwidth-\fboxsep-\fboxrule\relax}{\copyrighttext}}};
\end{tikzpicture}%
}
\begin{document}
\maketitle
\copyrightnotice
\begin{abstract}
   The International Fingerprint Liveness Detection Competition is an international biennial competition open to academia and industry with the aim to assess and report advances in Fingerprint Presentation Attack Detection. The proposed ``Liveness Detection in Action'' and ``Fingerprint representation'' challenges were aimed to evaluate the impact of a PAD embedded into a verification system, and the effectiveness and compactness of feature sets for mobile applications. Furthermore, we experimented a new spoof fabrication method that has particularly affected the final results. Twenty-three algorithms were submitted to the competition, the maximum number ever achieved by LivDet.
\end{abstract}
\section{Introduction}

Equipping a fingerprint sensor or verification system with Presentation Attacks Detection (PAD) allows handling a threat much more serious than that represented by zero-effort impostors \cite{marcel2019handbook}. The term PAD is defined by the ISO / IEC 30107 standard \cite{iso}, and in its most general definition, refers to all approaches aimed to identify false or altered characteristics in the biometric trait. Since these characteristics can come from artificial or dead fingerprints, these techniques were early referred to as ``vitality''/``liveness''  detection or ``antispoofing'' techniques \cite{chugh2018fingerprint,Marasco:2014:SAS:2658850.2617756,cnn}.

From 2006, once first specific approaches failed, software-based PAD methods followed the innovations attained in the pattern recognition research \cite{bushsurvey, Marasco:2014:SAS:2658850.2617756}. The increase of available data sets followed as well. Those methods relied on wavelet and textural features especially. More recently, the novel machine learning techniques inherited the idea coming from textural descriptors that filters adopted for feature extraction can be learnt more efficiently by convolutional neural networks \cite{cnn}; moreover, approaches outcoming the ``embeddings'' as final features were derived by appropriate deep learning architectures \cite{chugh2018fingerprint}. Finally, recent works underlined the need to assess the impact of PADs embedded into verification systems \cite{Chingovska2019}.
 
Over such years, the International Fingerprint Liveness Detection Competition (LivDet) \cite{8987281} allowed to make the point on the state-of-the-art of the fingerprint PADs performance defining common experimental protocols and data sets.
In this paper, we report the results achieved in the seventh LivDet edition. In LivDet 2019 \cite{8987281}, the competitors were asked to present algorithms outcoming an ``integrated score'', aimed to evaluate in only one measurement both liveness and matching scores.
Accordingly, we launched two different challenges in the 2021 edition: (1) Challenge 1 investigated how the integration of a PAD can impact the whole fingerprint verification performance; (2) Challenge 2 was devoted to the effectiveness assessment of feature vectors in terms of accuracy and compactness.

Spoofs (or Presentation Attack Instruments) of LivDet 2021 were collected by the well-known consensual methodology and distributed to the participants for the training step. The test set was made up of images from different people and materials, as usual, always consensually fabricated. However, we adopted an additional test set made up of semi-consensual replicas, acquired through the ScreenSpoof technique recently proposed in \cite{screenspoof}. This fake fabrication method turned out to be highly dangerous, as it does not require the full cooperation of the user. 
Therefore, we tested the competitors' algorithms even on this data set, and got interesting insights about the representativeness problem in data learning. We believe that the ScreenSpoof data set can become a good benchmark for assessing the PAD performance under more realistic attacks than those obtained with the user's consensus.




\section{Experimental Protocol and Evaluation}
\label{sec:Protocol}

\begin{figure}[!tbp]
  \centering
  \subfloat[GB CC]{\includegraphics[width=0.11\textwidth]{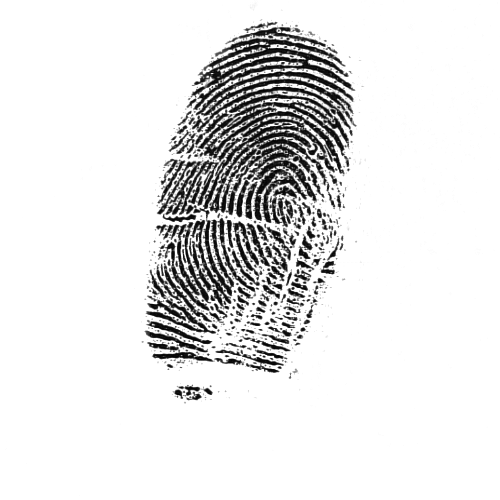}\label{fig:gbcc}}
  \subfloat[DL CC]{\includegraphics[width=0.11\textwidth]{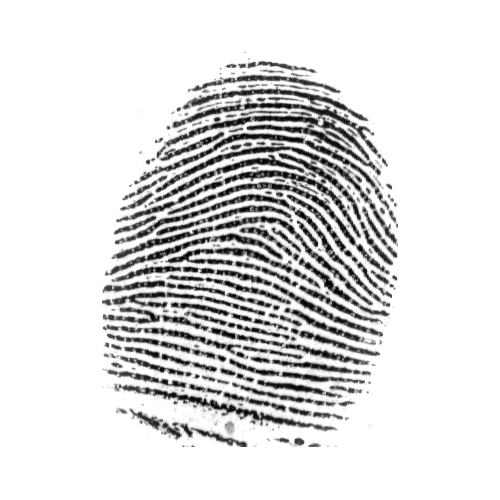}\label{fig:dlcc}}
  \subfloat[GB SS]{\includegraphics[width=0.11\textwidth]{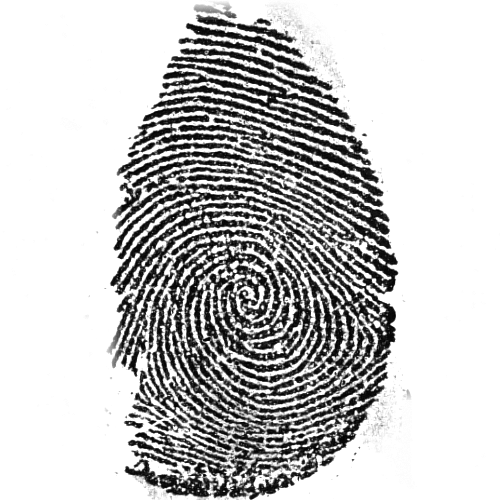}\label{fig:gbss}}
  \subfloat[DL SS]{\includegraphics[width=0.11\textwidth]{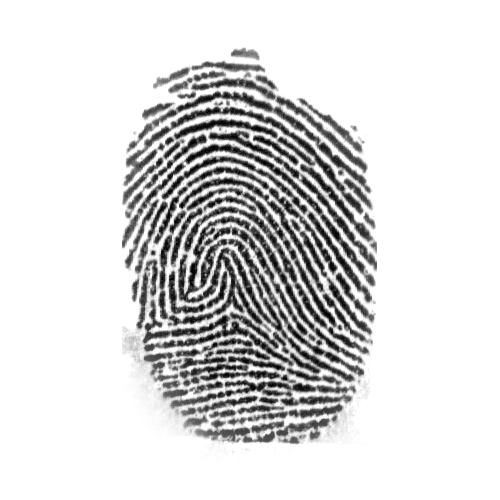}\label{fig:dlss}}
  \caption{Examples of acquired fake fingerprints. Consensual: GreenBit (a); Dermalog (b). ScreenSpoof: GreenBit (c); Dermalog (d).}
\end{figure}

Two distinct challenges characterize this competition:
\begin{itemize}
    \itemsep0em 
    \item Challenge 1, \textit{Liveness Detection in Action} \cite{8987281}: similarly to the previous edition, competitors were invited to submit a complete algorithm able to output both the probability of the image vitality (the so-called “liveness score”) given the extracted set of features and an integrated match score (“integrated score”) which includes the probability above with the probability of belonging to the claimed user. For this challenge, participants can decide whether to exploit the related “user-speciﬁc” information  \cite{userspec}).
    
    \item Challenge 2, \textit{Fingerprint representation}: In modern biometric systems, the feature vectors compactness and discriminability are fundamental to guarantee high performance in terms of accuracy and speed, especially for systems embedded in mobile devices. For this reason, we asked competitors to submit a liveness detection algorithm that returns, in addition to the probability of liveness, the feature vector corresponding to the input image.
    
\end{itemize}

\begin{table}[ht]
\centering
\caption{Participants name and submitted algorithms. The (*) symbol beside the name means that the participant used additional data to train the related model.}
\label{table:partecipantsLivDet}
\resizebox{0.7\textwidth}{!}{%
\begin{tabular}{|c|c|c|} 
\hline
\multicolumn{3}{|c|}{\textbf{Challenge 1}}                                                                                                              \\ 
\hhline{|===|}
\textbf{Participant}                                                                                         & \textbf{Algorithm name} & \textbf{Type}  \\ 
\hline
\multirow{2}{*}{Dermalog (*)}                                                                                & LivDet21CanC1           & Deep-learning  \\ 
\cline{2-3}
                                                                                                             & LivDet21DanC1           & Deep-learning  \\ 
\hline
\multirow{2}{*}{\begin{tabular}[c]{@{}c@{}}Hangzhou Jinglianwen \\Technology Co., Ltd.\end{tabular}}         & JLWLivDetW              & Hybrid         \\ 
\cline{2-3}
                                                                                                             & JLWLivDetD              & Hybrid         \\ 
\hline
\multirow{3}{*}{\begin{tabular}[c]{@{}c@{}}DIETI Università degli Studi di\\Napoli Federico II\end{tabular}} & ALD\_A                  & Deep-learning  \\ 
\cline{2-3}
                                                                                                             & ALD\_B                  & Deep-learning  \\ 
\cline{2-3}
                                                                                                             & ALD\_C                  & Deep-learning  \\ 
\hline
ZKTeco USA (*)                                                                                               & fingerprint\_antispoof  & Hybrid         \\ 
\hline
Hallym University                                                                                            & HallymMMC               & Deep-learning  \\ 
\hline
Brno University of Technology                                                                                & Damavand                & Hand-crafted   \\ 
\hhline{|===|}
\multicolumn{3}{|c|}{\textbf{Challenge 2}}                                                                                                              \\ 
\hhline{|===|}
\textbf{Participant}                                                                                         & \textbf{Algorithm name} & \textbf{Type}  \\ 
\hline
\multirow{2}{*}{Dermalog (*)}                                                                                & LivDet21ColC2           & Deep-learning  \\ 
\cline{2-3}
                                                                                                             & LivDet21DobC2           & Deep-learning  \\ 
\hline
LiveID Biometrics LLC                                                                                        & LiveID                  & Hybrid         \\ 
\hline
Unesp                                                                                                        & contreras               & Hand-crafted   \\ 
\hline
Anonymous$_{1}$                                                                                              & TYSYFingerNet           & Deep-learning  \\ 
\hline
\begin{tabular}[c]{@{}c@{}}Hangzhou Jinglianwen \\Tech. Co., Ltd.\end{tabular}                               & JLWLivDetL              & Hybrid         \\ 
\hline
\multirow{2}{*}{\begin{tabular}[c]{@{}c@{}}MEGVII (BEIJING) \\TECHNOLOGY CO, LTD\end{tabular}}               & megvii\_single          & Deep-learning  \\ 
\cline{2-3}
                                                                                                             & megvii\_ensemble        & Deep-learning  \\ 
\hline
\multirow{2}{*}{\begin{tabular}[c]{@{}c@{}}University of Applied \\Sciences Darmstadt\end{tabular}}          & hda                     & Deep-learning  \\ 
\cline{2-3}
                                                                                                             & PADUnk                  & Hand-crafted   \\ 
\hline
Chosun University                                                                                            & B\_ld2                  & Deep-learning  \\ 
\hline
\multirow{2}{*}{Anonymous$_{2}$}                                                                             & bb8                     & Hybrid         \\ 
\cline{2-3}
                                                                                                             & r2d2                    & Hybrid         \\
\hline
\end{tabular}}
\end{table}

\subsection{Data Sets}

\begin{table}[!t]
\caption{Device characteristics for LivDet2021 datasets.}
\label{table:sensors}
\centering
\resizebox{0.7\textwidth}{!}{%
\begin{tabular}[t]{ | l | l | c | c | c | c |}
\hline
\textbf{Scanner} & \textbf{Model} & \textbf{Res.[dpi]} & \textbf{Img Size} & \textbf{Format} &\textbf{Type} \\ \hline
Green Bit & DactyScan84C & 500 & 500x500 & BMP & Optical \\ \hline
Dermalog & LF10 & 500 &500x500 & PNG & Optical \\ \hline
\end{tabular}}
\end{table}

\begin{table*}[t]
\caption{Number of samples for each scanner and each part of the dataset.}
\label{table:datasetComposition}
\centering
\resizebox{\textwidth}{!}{%
\begin{tabular}{|c||c|c|c||c|c|c|c|c|c|}
\hline
& \multicolumn{3}{c||}{\textbf{Training}}      & \multicolumn{6}{c||}{\textbf{Test Consensual/ScreenSpoof}}  \\ \hline
\textbf{Dataset} & Live & Latex & RProFast & Live & Mix 1 & BodyDouble & ElmersGlue & GLS20 & RFast30\\ \hline
Green Bit       &1250&750&750&2050/2050&820/820&820/820&820/820&-&-\\ \hline
Dermalog      &1250&750&750&2050/2050&-&-&-&1230/1230&1230/1230\\ \hline

\end{tabular}%

}
\end{table*}

After the registration phase, each participant must sign a license agreement detailing the proper usage of data released for the competition. We reported in Table  \ref{table:partecipantsLivDet} the competitor and the correspondent algorithm names, the type of the solution presented, and the adherence details to the challenges. Although it was recommended only to use the LivDet training set, we also accepted solutions that used additional data. Since these competitors are in better condition than the others, this detail, shown in Tab. \ref{table:partecipantsLivDet}, has been considered in the analysis of the results.

The data sets for LivDet 2021 were collected from two different scanners: Green Bit DactyScan84C and Dermalog LF10. The devices are similar in image dimensions and resolution, as shown in Table \ref{table:sensors}.
For each sensor, we collected 11770 images. Live images came from multiple acquisitions of all fingers of different subjects.

One of the novelties of this edition concerns the spoof fabrication method. In addition to the classic consensual method, in which the mold is created through user collaboration, we have also employed the semi-consensual ScreenSpoof technique \cite{screenspoof}.
This method consists of taking a snapshot of the latent fingerprints left on a smartphone screen, imprinted during its regular use. After an appropriate preprocessing, aimed at segmenting and enhancing the fingerprint, the resulting ``negative'' image is printed on a transparent sheet and used as a mold, as in a traditional non-consensual process.

The entire datasets were divided into two parts by using images from different subjects: a training set for the configuration of the algorithms, composed by 25 users, and a test set, to evaluate the performance, composed by 41 users. 
The training set, released to participants, comprises spoofs acquired only with the consensual method and fabricated using different materials from those used in the test set since the lack of awareness of spoofing type allows representing a real attack scenario.
The unknown materials used for this edition are GLS20 and RFast30 for Dermalog, while for GreenBit, we used Body Double, ElmersGlue, and Mix 1, a new material introduced in the previous edition. Further details on dataset composition are reported in Table \ref{table:datasetComposition}.
Moreover, to further assess algorithms' reliability, we inserted fingerprint samples from both consensual and ScreenSpoof techniques in the test set. Thus, we designed cross-material and cross-method experiments, following the typical fingerprint PAD evaluation protocols.

\subsection{Algorithms Submission}

The algorithm submission for Challenge 1 uses the same structure as LivDet 2019. Each submitted algorithm is a console application with the following list of parameters:
\\
\textit{[nameOfAlgorithm] [ndataset] [templateimagesfile] [probeimagesfile][livenessoutputfile] [IMSoutputfile]}
\\The parameter \textit{[ndataset]} is the dataset identification number, \textit{ [templateimagesfile]} is the text file name with the list of absolute paths of each template image registered in the system, while  \textit{ [probeimagesfile]} is the text file with the list of absolute paths of each image to analyse. The last two parameters are the path of the output files where the algorithm saves the result regarding every probe image. In the \textit{[livenessoutputfile]} the liveness output of each processed image is saved. This output is the degree of ``liveness'' normalized in the range 0 and 100 (100 is the maximum degree of liveness, 0 means that the image is fake). Scores [0, 50) classify fingerprint image as ``fake'' while scores [50,100] classify fingerprint image as ``live''. The \textit{[IMSoutputfile]} reports for each probe the combined probability of being a live fingerprint and belonging to the declared identity, normalized in the range 0 and 100.  An IMSoutput between [0, 50) classify fingerprint image as ``fake'' or belonging to an attacker, while a value between [50,100] classify fingerprint image as ``live'' and belonging to the declared user. The selected classification threshold in order to measure the performance is 50. In both outputs, if the algorithm has not been able to process the image, the corresponding value will be -1000. 

The algorithms submission process for Challenge 2 is different from that of all previous editions. Each submitted algorithm is a console application with the following list of parameters: \\\textit{[nameOfAlgorithm].exe [ndataset] [probeimagesfile][livenessoutputfile] [embeddingsfile]}\\
The parameters \textit{[ndataset],[probeimagesfile]} and \textit{[livenessoutputfile]} observe the same convention as listed above while \textit{[embeddingsfile]}, a LivDet 2021 novelty, represents the file of the feature vectors of each processed image.

The two challenges' operating requirements were communicated to the participants before the competition start.
\subsection{Performance Evaluation}
\label{sec:protocol}
In both challenges the performance of the FPADs will be evaluated using three basic PAD metrics:
    \begin{itemize}
    \itemsep0em 
    \item Liveness Accuracy: percentages of samples correctly classified by the PAD.
    \item BPCER (Bona fide Presentation Classification Error Rate): Rate of misclassified live fingerprints \cite{iso}.
    \item APCER (Attack Presentation Classification Error Rate): Rate of misclassified fake fingerprints \cite{iso}.
    \end{itemize}
In Challenge 1, in addition to the liveness detection accuracy, the performance of the integrated system is evaluated with the following metrics:
\begin{itemize}
    \itemsep0em 
    \item FNMR (False Non-Match Rate): Rate of genuine live fingerprints classified as impostor.
    \item FMR (False Match Rate): Rate of zero-effort impostors classified as genuine.
    \item IAPMR (Impostor Attack Presentation Match Rate): rate of impostor attack presentations classified as genuine.
    \item Integrated Matching (IM) Accuracy: percentages of samples correctly classified by the integrated system.
\end{itemize}
All the comparisons were made with the templates belonging to live fingerprints, while the probes could be both live or fake in order to simulate a real scenario.
Templates can be compared with an image of the live fingerprint corresponding to the same user and the same finger (genuine), with an image of the fake fingerprint corresponding to the same user and the same finger (presentation attack) or with an image of the live fingerprint of another user (zero-effort  attack).
The number of comparisons made for each part of the dataset is 8,200 for genuine users and presentation attacks, 8,000 for zero-effort impostors.
It is worth mentioning that, although a fingerprint presentation attack using templates belonging to a fake sample cannot be excluded, we have not taken into consideration this particular case.


In Challenge 2, we evaluated the compactness and representativeness of the feature vectors. Since it was possible to submit algorithms on different operating systems, we considered a comparison of computational times to be unfair and we preferred to make a comparison on the size of the feature vectors generated by the systems. For this reason, we have reported the final dimensions in terms of the size of the single feature vector. We have also trained and tested an SVM by designing the following two tasks: 
\begin{itemize}
    \item Client task: we selected live and fake images from 11 random users of the test set (25\% of test users) to train the SVM and the remaining 75\% of test users to test it. We repeated this experiment ten times and computed the mean and standard deviation for each dataset.
    \item Material task: for each sensor, we composed the SVM training set by merging all samples from a specific material and an equal number of live samples, randomly selected from all users. We tested on remaining test data. We repeated the experiment for each material.
    
\end{itemize}

\section{Discussions on the reported results}
\label{sec:Result}

This section discusses the performance of the submitted algorithms to the two LivDet2021 challenges.

\begin{figure}[ht]

  \centering
  \subfloat[BD]{\includegraphics[width=0.10\textwidth]{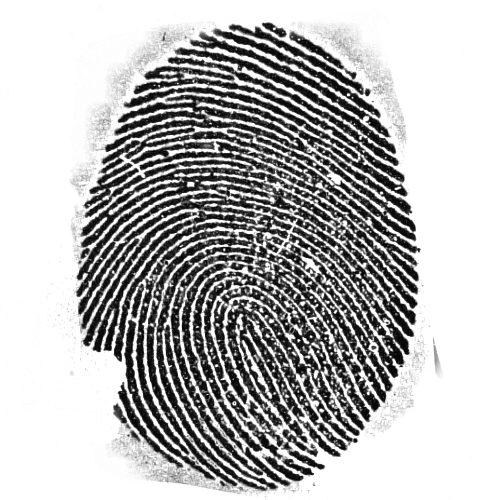}\label{fig:bd}}
  \subfloat[GLS20]{\includegraphics[width=0.11\textwidth]{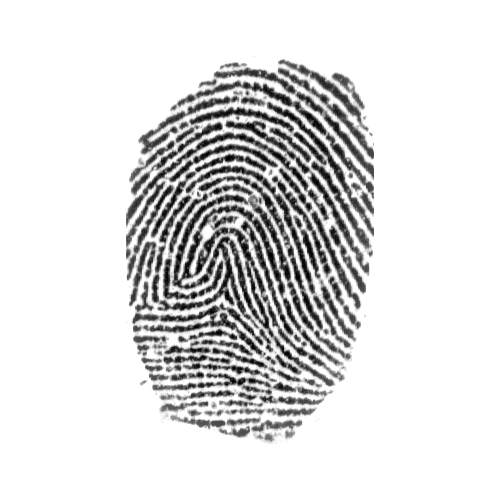}\label{fig:gls}}
  \subfloat[RF30]{\includegraphics[width=0.11\textwidth]{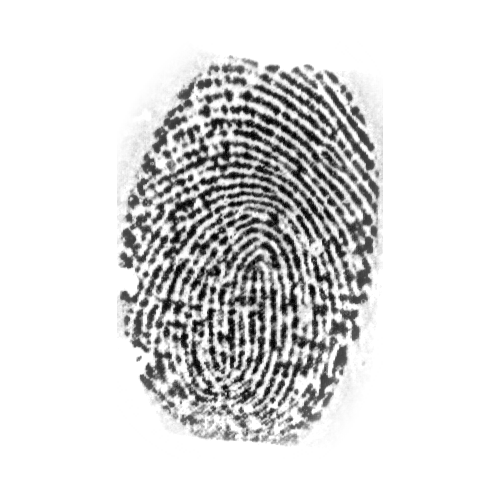}\label{fig:ddddss}}
  \caption{Examples of fake fingerprints composed of different materials erroneously classified as live by all the PADs.}
  \label{fig:allfake}
\end{figure}

Challenge 1 aims to investigate the integration of PADs into real verification systems.
The results are summarized in Table \ref{tab:IMS_results}. 
The first important consideration concerns the difference in performance between consensus-based spoofs (CC) and screenspoofs (SS).
Since the training set is acquired by the consensual method, SS data represents some sorts of ``advanced'' never-seen-before attacks. We reported in Fig. \ref{fig:allfake} some examples of spoofs misclassified by all detectors, to show the high quality of the fakes, including the presence of pores, which justifies the wrong classification. Accordingly, the PADs performance is notably lower in this case.

It is worth noting that, regardless of the dataset, a higher IM accuracy always corresponds to higher liveness accuracy. This means the PAD system greatly influences the performance of the integrated system, whatever fusion rules and matching algorithms.  
Tab. \ref{fig:ch1over} reports the average IM accuracy on the four datasets: the algorithm that obtains the best performance is LivDet21CanC1, which is the one that is least affected by the advanced never-seen-before attacks above. However, it is important to point out that LivDet21CanC1 was trained on additional data with respect to the ones made available for this competition. Among the algorithms trained on the LivDet2021 training set only, the best method is ALD\_C (Adversarial Liveness Detection) which leverages adversarial perturbations as a way to perform a wide and targeted data  augmentation. It is important to highlight that some algorithms, such as fingerprint\_antispoof, have a high imbalance between FNMR and FMR, probably due to an incorrect parameterization.
\begin{table}[h]
\caption{Challenge 1 Integrated and Liveness results.}
\label{tab:IMS_results}
\centering
\resizebox{0.7\textwidth}{!}{%
\begin{tabular}{|c|c||c|c|c|c||c|c|c|}
\hline
 &   \textbf{Algorithms}           & \textbf{FNMR}  & \textbf{FMR}   & \textbf{IAPMR} & \textbf{\begin{tabular}[c]{@{}c@{}}IM\\Acc.\end{tabular}}& \textbf{BPCER}& \textbf{APCER}& \textbf{\begin{tabular}[c]{@{}c@{}}Liv\\Acc.\end{tabular}} \\ \hline
\multirow{10}{*}{\begin{sideways}\textbf{GreenBit  CC}\end{sideways}} & LivDet21CanC1      & 0.84  & 0.29  & 12.89 & 95.29 & 0.14  & 29.98 & 89.83\\      
 & LivDet21DanC1      & 1.15  & 0.29  & 11.13 & 95.78 & 0.45  & 25.42 & 91.16\\           
 & hallymMMC              & 53.43 & 0.80  & 0.01  & 81.78 & 1.60  & 39.17 & 85.77\\         
 & fingerprint\_antispoof & 1.16  & 98.51 & 13.80 & 62.67 & 0.61  & 44.82 & 84.53\\          
 & JLWLivDetW             & 2.38  & 1.24  & 4.84  & \textbf{97.17} & 1.70  & 8.16  & 96.13\\    
 & JLWLivDetD             & 2.77  & 3.93  & 6.26  & 95.68 & 1.70  & 8.16  & 96.13\\           
 & ALD\_A                 & 68.51 & 31.24 & 31.11 & 56.28 & 0.35  & 23.29 & 91.93\\         
 &ALD\_B                 & 8.45  & 0.19  & 1.43  & 96.62 & 0.40  & 28.37 & 90.20\\          
 &ALD\_C                 & 8.40  & 0.19  & 1.50  & 96.61 & 0.43  & 31.82 & 89.36\\          
 &Damavand                   & 47.71 & 57.26 & 26.32 & 56.35 & 45.25 & 27.12 & 60.80\\\hline
\multirow{10}{*}{\begin{sideways}\textbf{Dermalog  CC}\end{sideways}}  & LivDet21CanC1      & 2.78  & 0.26  & 0.09  & 98.95 & 1.56  & 0.18  & 98.90\\        
 & LivDet21DanC1      & 2.24  & 0.24  & 0.17  & \textbf{99.11} & 1.23  & 0.28  & 99.09\\   
 & hallymMMC              & 49.06 & 0.93  & 0.05  & 83.19 & 1.81  & 7.74  & 96.19\\          
 & fingerprint\_antispoof & 1.27  & 98.18 & 15.00 & 62.34 & 0.31  & 23.50 & 91.90\\          
 & JLWLivDetW             & 1.10  & 1.84  & 2.52  & 98.18 & 0.35  & 2.84  & 98.81\\          
 & JLWLivDetD             & 0.96  & 4.94  & 2.65  & 97.17 & 0.35  & 2.84  & 98.81\\          
 & ALD\_A                 & 11.88 & 0.00  & 1.37  & 95.55 & 0.15  & 3.30  & 98.79\\          
 & ALD\_B                 & 11.84 & 0.00  & 2.21  & 95.28 & 0.12  & 4.85  & 98.29\\          
 & ALD\_C                 & 12.00 & 0.00  & 2.10  & 95.26 & 0.26  & 4.74  & 98.23\\          
 & Damavand                   & 34.71 & 60.95 & 21.38 & 61.17 & 36.85 & 21.38 & 68.35\\\hline
\multirow{10}{*}{\begin{sideways}\textbf{GreenBit  SS}\end{sideways}} & LivDet21CanC1      & 0.65  & 0.21  & 15.52 & 94.50 & 0.07  & 25.00 & 91.55\\          
 & LivDet21DanC1      & 0.96  & 0.21  & 1.89  &\textbf{ 98.97} & 0.39  & 3.28  & 98.64\\          
 & hallymMMC              & 52.60 & 0.53  & 0.07  & 82.13 & 1.66  & 75.95 & 73.37\\          
 & fingerprint\_antispoof & 1.30  & 98.39 & 32.78 & 56.29 & 0.64  & 62.33 & 78.63\\          
 & JLWLivDetW             & 2.38  & 1.19  & 35.43 & 86.91 & 1.72  & 54.44 & 80.57\\          
 & JLWLivDetD             & 2.65  & 3.98  & 37.54 & 85.19 & 1.72  & 54.44 & 80.57\\          
 & ALD\_A                 & 62.99 & 37.24 & 37.18 & 54.13 & 0.38  & 81.01 & 72.52\\          
 & ALD\_B                 & 8.52  & 0.03  & 20.44 & 90.26 & 0.41  & 82.31 & 72.21\\          
 & ALD\_C                 & 8.50  & 0.03  & 20.39 & 90.28 & 0.45  & 83.28 & 71.71\\          
 & Damavand                   & 35.89 & 59.50 & 19.50 & 61.88 & 38.17 & 20.62 & 67.73\\\hline
\multirow{10}{*}{\begin{sideways}\textbf{Dermalog SS}\end{sideways}}  & LivDet21CanC1      & 2.80  & 0.34  & 33.79 & 87.59 & 1.44  & 59.06 & 79.19\\          
 & LivDet21DanC1      & 2.26  & 0.34  & 23.56 & \textbf{91.21} & 1.20  & 31.60 & 88.59\\         
 & hallymMMC              & 49.06 & 0.85  & 0.30  & 83.13 & 1.70  & 26.56 & 89.94\\          
 & fingerprint\_antispoof & 1.20  & 98.66 & 37.33 & 54.70 & 0.34  & 74.29 & 74.81\\          
 & JLWLivDetW             & 1.12  & 1.83  & 74.38 & 40.80 & 0.37  & 40.24 & 66.35\\          
 & JLWLivDetD             & 1.06  & 4.98  & 75.70 & 72.57 & 0.37  & 40.24 & 66.35\\          
 & ALD\_A                 & 11.96 & 0.00  & 20.67 & 89.03 & 0.12  & 77.06 & 74.02\\          
 & ALD\_B                 & 11.94 & 0.00  & 21.51 & 88.76 & 0.09  & 66.43 & 77.61\\          
 & ALD\_C                 & 12.10 & 0.00  & 21.15 & 88.83 & 0.23  & 60.09 & 79.65\\         
 & Damavand                   & 33.39 & 56.78 & 63.55 & 48.81 & 38.25 & 65.28 & 52.69\\\hline
\end{tabular}}
\end{table}

\begin{table}[]
\caption{Ch.1 IM average accuracy over the 4 datasets.}
\label{fig:ch1over}
\centering
\resizebox{0.5\textwidth}{!}{%
\begin{tabular}{|l|c|}
\hline
\textbf{Algorithm}     & \textbf{\begin{tabular}[c]{@{}c@{}}Overall IM Acc. [\%]\end{tabular}} \\ \hline
LivDet21CanC1          & 94.08    \\ \hline
LivDet21DanC1          & \textbf{96.27}                         \\ \hline
hallymMMC              & 82.56    \\ \hline
fingerprint\_antispoof & 59.00    \\ \hline
JLWLivDetW             & 80.76    \\ \hline
JLWLivDetD             & 87.65    \\ \hline
ALD\_A                 & 73.75    \\ \hline
ALD\_B                 & 92.73    \\ \hline
ALD\_C                 & 92.75    \\ \hline
Damavand                   & 57.05    \\ \hline
\end{tabular}}
\end{table}

Challenge 2 adds as evaluation parameter the compactness and representativeness of the feature vectors used.
Tab. \ref{tab:overalaccuracych2} show the systems performance. 
It is possible to notice a general imbalance between BPCER and APCER. The latter is, in fact, higher than the former, with a few exceptions. This imbalance has already been highlighted in previous editions of LivDet \cite{8987281}, and is probably due to the use of never-seen-before materials in the test set. The method with the greatest overall accuracy is megvii\_ensemble with 93.79\%.
The general results are in line with Challenge 1, although these results are not directly comparable as the experimental protocol is different.
Tab.s \ref{tab:clientsvmaccuracych2}-\ref{tab:materialsvmch2} show the results for the client task and material task experiments, respectively, whose experimental protocol is reported in Sec. \ref{sec:protocol}.
A higher average accuracy characterizes these experiments. In fact, the client task does not contain never-seen-before materials. On the other hand, in the material task, training and test sets share the same users, and the user-specific effect allows higher accuracy than in previous experiments and confirms what reported in \cite{userspec}.
Tab. \ref{tab:clientsvmaccuracych2} shows the feature vectors size of each algorithm. The most representative feature vectors in terms of accuracy on the two tasks are those of megvii-ensemble with a size of 256 features. It is interesting to note the high performance obtained by the two handcrafted features-based systems of the challenge, contreras and PADUnk. Unfortunately, PADUnk lacks compactness and has the largest-sized feature vector. This is common in many handcrafted features-based PADs. The most compact feature vector is that of megvii-single which with 64 components, and is always in the top-3 rank for all tasks.

\begin{table*}[ht]
\caption{Challenge 2 Liveness accuracy of the algorithms on the test sets. For each dataset  the rate of misclassified live  and fake fingerprints are reported. The last column is relative to the average of the total accuracy on the four datasets.}
\label{tab:overalaccuracych2}
\centering
\resizebox{\textwidth}{!}{%
\begin{tabular}{|l||c|c|c|c|c||c|c|c|c|c||c|} 
\hline
\multicolumn{1}{|c||}{\multirow{3}{*}{\textbf{Algorithm} }} & \multicolumn{5}{c||}{\textbf{Green Bit~} }                  & \multicolumn{5}{c||}{\textbf{Dermalog} }                      & \multirow{3}{*}{\begin{tabular}[c]{@{}c@{}}\textbf{Overall}\\\textbf{Liveness}\\\textbf{Acc.[\%]} \end{tabular}}  \\ 
\cline{2-11}
\multicolumn{1}{|c||}{}       & \multicolumn{1}{c|}{\multirow{2}{*}{\begin{tabular}[c]{@{}c@{}} \textit{BPCER}\\\textit{ [\%]} \end{tabular}}} & \multicolumn{2}{c|}{CC}          & \multicolumn{2}{c||}{SS}            & \multicolumn{1}{c|}{\multirow{2}{*}{\begin{tabular}[c]{@{}c@{}} \textit{BPCER}\\\textit{ [\%]} \end{tabular}}} & \multicolumn{2}{c|}{CC}                     & \multicolumn{2}{c||}{SS}   &    \\ 
\cline{3-6}\cline{8-11}
\multicolumn{1}{|c||}{}       & \multicolumn{1}{c|}{}                              & \begin{tabular}[c]{@{}c@{}}\textit{APCER}\\\textit{ [\%]} \end{tabular} & \begin{tabular}[c]{@{}c@{}}\textit{Liveness }\\\textit{ Acc.[\%]} \end{tabular} & \begin{tabular}[c]{@{}c@{}}\textit{APCER}\\\textit{ [\%]}\end{tabular} & \begin{tabular}[c]{@{}c@{}}\textit{Liveness}\\\textit{ Acc.[\%]}\end{tabular} & \multicolumn{1}{c|}{}& \begin{tabular}[c]{@{}c@{}}\textit{APCER}\\\textit{ [\%]} \end{tabular} & \begin{tabular}[c]{@{}c@{}}\textit{Liveness}\\\textit{ Acc.[\%]} \end{tabular} & \begin{tabular}[c]{@{}c@{}}\textit{APCER}\\\textit{ [\%]} \end{tabular} & \begin{tabular}[c]{@{}c@{}}\textit{Liveness }\\\textit{ Acc.[\%]} \end{tabular} &                           \\ 
\hline
LiveID                        & 15.76                & 40.98       & 70.49               & 11.63      & 86.50             & 15.27                & 26.71       & 78.49              & 33.66       & 74.7                & 77.55                   \\
LivDetColC2                   & 0.20                 & 29.88       & 83.61               & 24.76      & 86.41             & 1.61                 & 0.16        & 99.18              & 58.86       & 67.16               & 59.34                   \\
LivDetDobC2                   & 0.59                 & 25.41       & 85.88               & 3.25       & \textbf{97.96}             & 1.07                 & 0.28        & 99.37              & 31.34       & 82.41               & 91.41                   \\
contreras                     & 8.98                      & 3.94        & 93.77               & 26.67      & 81.37             & 5.27                 & 7.60        & 93.46              & 73.94       & 57.27               & 81.47                   \\
TYSYFingerNet                 & 13.12                & 54.19       & 64.48               & 58.38      & 62.19             & 2.10                 & 30.65       & 82.33              & 79.31       & 55.79               & 66.20                   \\
JLWLivDetL                    & 2.59                 & 8.21        & 94.35               & 54.11      & 69.31             & 0.68                 & 2.80        & 98.16              & 95.12       & 45.41               & 76.81                   \\
megvii\_single                 & 0.29                 & 6.30        & 96.43               & 13.94      & 92.26             & 0.83                 & 0.77        & 99.20              & 29.07       & 83.77               & 92.92                   \\
megvii\_ensemble               & 0.05                 & 2.72        & \textbf{98.49}               & 13.62      & 92.55             & 0.24                 & 0.04        & \textbf{99.87}              & 28.66       & 84.26               & \textbf{93.79}                   \\
hda                           & 4.10                 & 24.11       & 84.99               & 81.71      & 53.55             & 7.95                 & 13.13       & 89.22              & 58.46       & 64.57               & 73.08                   \\
PADUnk                        & 1.46                 & 37.20       & 79.05               & 18.42      & 89.29             & 2.68                 & 4.80        & 96.16              & 24.72       & \textbf{85.30}               & 87.45                   \\
B\_d2                         & 3.61                 & 5.37        & 95.43               & 27.56      & 83.32             & 2.59                 & 8.33        & 94.28              & 77.97       & 56.30               & 82.33                   \\
bb8                           & 3.46                 & 7.85        & 94.15               & 39.80      & 76.72             & 2.39                 & 4.27        & 96.58              & 69.51       & 46.03               & 78.37                   \\
r2d2                          & 2.20                 & 12.36       & 92.26               & 57.93      & 67.06             & 1.27                 & 2.56        & 98.03              & 82.11       & 45.85               & 75.80                   \\
\hline
\end{tabular}
}
\end{table*}

\begin{table*}[]
\caption{Challenge 2 Client Task accuracy: a linear SVM is trained on 11 random users of the test set (25\% of users) and tested on the remaining 30 users.}
\label{tab:clientsvmaccuracych2}
\centering
\resizebox{\textwidth}{!}{%
\begin{tabular}{|l|c|c|c|c|c|} 
\hline
\multicolumn{1}{|c|}{\multirow{2}{*}{\textbf{Algorithm} }} & \multirow{2}{*}{\begin{tabular}[c]{@{}c@{}}\textbf{Feature}\\\textbf{vect. size}\end{tabular}} & \multicolumn{4}{c|}{\textbf{Linear SVM Accuracy [\%] - Client Task} }                          \\ 
\cline{3-6}
\multicolumn{1}{|c|}{}       &        & \textbf{GreenBit CC}  & \textbf{Dermalog CC}  & \textbf{GreenBit SS}  & \textbf{Dermalog SS}   \\ 
\hline
LiveID                       & 98     & $94.05 \pm 1.91$      & $94.16 \pm 0.87$      & $95.50 \pm 1.33$      & $97.53 \pm 0.90$        \\ 
\hline
LivDet21ColC2                & 128    & $99.07 \pm 0.24$      & $98.92 \pm 0.25$      & $85.20 \pm 18.24$     & $95.24 \pm 1.52$       \\ 
\hline
LivDet21DobC2                & 128    & $99.14 \pm 0.31$      & $99.09 \pm 0.18$      & $85.49 \pm 18.41$     & $97.65 \pm 0.51$       \\ 
\hline
contreras                    & 1062   & $95.53 \pm 1.19$      & $97.52 \pm 0.82$      & $97.57 \pm 0.95$      & $\bm{99.41 \pm 0.54}$       \\ 
\hline
TYSYFingerNet                & 256    & $90.40 \pm 1.43$       & $92.29 \pm 0.53$      & $67.96 \pm 1.64$      & $87.47 \pm 1.14$       \\ 
\hline
JLWLivDetL                   & 1280   & $98.00 \pm 0.94$      & $98.85 \pm 0.61$      & $97.5 \pm 1.60$        & $96.93 \pm 1.44$       \\ 
\hline
megvii\_single                             & 64     & $99.32 \pm 0.27$      & $99.36 \pm 0.19$      & $96.86 \pm 1.20$       & $97.51 \pm 0.62$       \\ 
\hline
megvii\_ensemble                           & 256    & $\bm{99.61 \pm 0.22}$    & $\bm{99.75 \pm 0.13}$      & $\bm{98.40 \pm 0.78}$       & $98.27 \pm 0.50$        \\ 
\hline
hda                          & 1280   & $95.06 \pm 1.16$      & $93.80 \pm 0.76$       & $94.32 \pm 2.12$      & $96.26 \pm 1.44$       \\ 
\hline
PADUnkEnFv                   & 65536                                & $96.33 \pm 1.27$      & $96.08 \pm 0.95$      & $98.18 \pm 1.29$      & $99.34 \pm 0.72$       \\ 
\hline
B\_d                         & 4096   & $94.95 \pm 1.14$      & $94.83 \pm 0.52$      & $90.05 \pm 1.25$      & $91.63 \pm 1.61$       \\ 
\hline
bb8                          & 192    & $96.39 \pm 0.53$      & $97.98 \pm 0.33$      & $90.44 \pm 1.05$      & $90.67 \pm 1.33$       \\ 
\hline
r2d2                         & 192    & $96.84 \pm 0.85 $     & $99.08 \pm 0.39 $     & $94.53 \pm 1.66$      & $82.67 \pm 2.97$       \\
\hline
\end{tabular} }
\end{table*}

\begin{table*}[]
\caption{Challenge 2 Material Task accuracy: a linear SVM is trained on all samples coming from a specific material and an equal number of live samples and tested on the remaining test data. For example BD acc. is the liveness accuracy of a SVM trained on BodyDouble and tested on Mix 1 and ElmersGlue for the Green Bit sensor.}
\label{tab:materialsvmch2}
\centering
\resizebox{\textwidth}{!}{%
\begin{tabular}{|l|c|c|c|c|c|c|c|c|c|c|c|c|c|c|}
\hline
\multicolumn{1}{|c|}{\multirow{4}{*}{\textbf{Algorithm}}} & \multicolumn{14}{c|}{\textbf{Linear SVM Accuracy {[}\%{]} - Material task}}                      \\ \cline{2-15} 
\multicolumn{1}{|c|}{}      & \multicolumn{4}{c|}{\textbf{Green Bit CC}}              & \multicolumn{3}{c|}{\textbf{Dermalog CC}}                 & \multicolumn{4}{c|}{\textbf{Green Bit SS}}              & \multicolumn{3}{c|}{\textbf{Dermalog SS}}                          \\ \cline{2-15} 
\multicolumn{1}{|c|}{}      & \multirow{2}{*}{\textbf{\begin{tabular}[c]{@{}c@{}}BD\\ Acc.\end{tabular}}} & \multirow{2}{*}{\textbf{\begin{tabular}[c]{@{}c@{}}Mix1\\ Acc.\end{tabular}}} & \multirow{2}{*}{\textbf{\begin{tabular}[c]{@{}c@{}}EGlue\\ Acc.\end{tabular}}} & \multirow{2}{*}{\textit{\textbf{\begin{tabular}[c]{@{}c@{}}Mean\\ Acc.\end{tabular}}}} & \multirow{2}{*}{\textbf{\begin{tabular}[c]{@{}c@{}}GLS\\ Acc.\end{tabular}}} & \multirow{2}{*}{\textbf{\begin{tabular}[c]{@{}c@{}}RFast\\ Acc.\end{tabular}}} & \multirow{2}{*}{\textbf{\begin{tabular}[c]{@{}c@{}}Mean\\ Acc.\end{tabular}}} & \multirow{2}{*}{\textbf{\begin{tabular}[c]{@{}c@{}}BD\\ Acc.\end{tabular}}} & \multirow{2}{*}{\textbf{\begin{tabular}[c]{@{}c@{}}Mix1\\ Acc.\end{tabular}}} & \multirow{2}{*}{\textbf{\begin{tabular}[c]{@{}c@{}}EGlue\\ Acc.\end{tabular}}} & \multirow{2}{*}{\textit{\textbf{\begin{tabular}[c]{@{}c@{}}Mean\\ Acc.\end{tabular}}}} & \multirow{2}{*}{\textbf{\begin{tabular}[c]{@{}c@{}}GLS\\ Acc.\end{tabular}}} & \multirow{2}{*}{\textbf{\begin{tabular}[c]{@{}c@{}}RFast\\ Acc.\end{tabular}}} & \multirow{2}{*}{\textit{\textbf{\begin{tabular}[c]{@{}c@{}}Mean\\ Acc.\end{tabular}}}} \\
\multicolumn{1}{|c|}{}      &                 &                   &                    &                            &                  &                    &                   &                 &                   &                    &                            &                  &                    &                            \\ \hline
LiveID                      & 61.92           & 78.47             & 84.74              & 75.04                      & 94.83            & 95.95              & 95.39             & 85.64           & 97.63             & 92.26              & 91.84                      & 97.8             & 98.73              & 98.27                      \\ \hline
LivDetColC2             & 98.43           & 99.44             & 99.16              & \textbf{99.01 }                     & 99.51            & 99.71              & 99.61             & 99.23           & 57.14             & 95.30               & 83.89                      & 98.93            & 97.71              & 98.32                      \\ \hline
LivDetDobC2             & 98.47           & 99.34             & 99.16              & 98.99                      & 99.32            & 99.61              & 99.47             & 99.20            & 57.14             & 97.91              & 84.75                      & 99.37            & 98.05              & 98.71                      \\ \hline
contreras                   & 77.15           & 89.79             & 93.76              & 86.90                       & 99.32            & 98.98              & 99.15             & 73.41           & 99.48           & 97.70               & 90.20                       & 99.8             & 99.95              & 99.88                      \\ \hline
TYSYFingerNet               & 48.85           & 84.49             & 79.97              & 71.10                       & 92.54            & 92.00                 & 92.27             & 58.05           & 58.08             & 45.61              & 53.91                      & 89.46            & 86.63              & 88.05                      \\ \hline
JLWLivDetL                  & 84.32           & 99.06             & 99.65              & 94.34                      & 99.71            & 99.76              & 99.74             & 91.53           & 99.37             & 96.55              & 95.82                      & 98.98            & 98.00                 & 98.49                      \\ \hline
megvii\_single              & 87.18           & 92.96             & 97.28              & 92.47                      & 99.61            & 99.41              & 99.51             & 94.29           & 97.84             & 93.80               & 95.31                      & 98.59            & 98.44              & 98.52                      \\ \hline
megvii\_ensemble            & 86.06           & 97.98             & 99.76              & 94.6                       & 99.85            & 99.76              &\textbf{ 99.81 }            & 96.52           & 98.78             & 97.18              & \textbf{97.49}                      & 99.51            & 99.66              & 99.59                      \\ \hline
hda                         & 67.94           & 84.04             & 85.12              & 79.03                      & 96.20             & 95.66              & 95.93             & 83.38           & 94.81             & 88.61              & 88.93                      & 97.22            & 98.98              & 98.10                       \\ \hline
PADUnk                      & 66.59           & 78.4              & 80.80               & 75.26                      & 98.54            & 99.22              & 98.88             & 81.08           & 99.27             & 92.79              & 91.05                      & 100.00              & 99.95              & \textbf{99.98}                      \\ \hline
B\_ld2                      & 86.52           & 95.47             & 96.93              & 92.97                      & 96.54            & 95.71              & 96.13             & 82.86           & 89.27             & 89.34              & 87.16                      & 94.49            & 95.37              & 94.93                      \\ \hline
bb8                         & 90.73           & 96.83             & 97.56              & 95.04                      & 98.54            & 98.24              & 98.39             & 89.55           & 90.38             & 88.92              & 89.62                      & 92.34            & 93.41              & 92.88                      \\ \hline
r2d2                        & 88.40            & 97.21             & 98.36              & 94.66                      & 99.66            & 98.63              & 99.15             & 95.57           & 96.72             & 94.18              & 95.49                      & 86.98            & 88.93              & 87.96                      \\ \hline
\end{tabular}}
\end{table*}

\section{Conclusions}
    
\label{sec:Conclusion}
The seventh edition of LivDet investigated the impact of integrating a PAD system with a fingerprint verification system and the level of compactness and representativeness achieved by feature sets adopted.

First of all, the results showed the strong influence that PADs have on the integrated results.
Furthermore, this edition highlighted that severe vulnerabilities are still present; the data availability is only a partial solution. Both integrated and PAD systems reported a notable, unexpected performance decrease for SS-based attacks. This adds a novel freedom degree for attackers: the fake fabrication method, which is part of a never-seen-before attack beside a novel material.
This can represent a serious threat, given that, for example, the ScreenSpoof is a relatively simple counterfeit method, but much more realistic than attacks by coercion, as the CC-based attacks.
Based on these results, integration of PAD and matching systems is worth investigations: reported accuracy is still far from that of the best off-the-shelf matching systems not equipped with liveness detection capability.

On the other hand, compactness and representativeness of current embeddings appeared promising: the best method of Challenge 2 achieved an overall liveness accuracy higher than 90\% using a feature vector of 64 components only. We believe that there is room for further improvements.

To sum up, the seventh edition of LivDet helped to point out the current state of PAD systems: handcrafted, deep learning and hybrid algorithms were presented, exhibiting different pros and cons. Deep-learning methods showed the best level of compactness and representativeness, while handcrafted algorithms reported a higher generalization ability, without the need for additional training data, particularly on the Dermalog SS test set. They did not seem, on average, fully effective when integrated into matching systems, but significant efforts have been made to improve the feature vector representativeness and the generalization ability. However, never-seen-before attacks may still be a crucial limitation in their successful adoption.

{\small
\bibliographystyle{abbrv}
\bibliography{bibliography}
}

\end{document}